\documentclass[letterpaper, 10 pt, conference]{ieeeconf}  

\IEEEoverridecommandlockouts                              

\usepackage{graphics} 
\usepackage{epsfig} 
\usepackage{mathptmx} 
\usepackage{times} 
\usepackage{amsmath} 
\usepackage{bm}
\usepackage{amssymb}  
\usepackage{newtxtext, newtxmath}
\usepackage{hyperref}

\usepackage{subcaption}
\usepackage{algorithmic}
\usepackage{booktabs}
\usepackage{adjustbox}
\usepackage{multirow}
\usepackage[lined,boxed,ruled,noend]{algorithm2e}
\usepackage{balance}

\usepackage{xcolor}

\newcommand{\abbv}{\texttt{CaDRE}\xspace}
\newcommand{\para}[1]{\noindent\textbf{#1.}}

\title{\LARGE \bf
CaDRE: Controllable and Diverse Generation of Safety-Critical Driving Scenarios using Real-World Trajectories
}

\author{Peide Huang$^{1}$ \quad Wenhao Ding$^{1}$ \quad Benjamin Stoler$^{1}$ \quad Jonathan Francis$^{1,2}$ \quad Bingqing Chen$^{2}$  \quad Ding Zhao$^{1}$
\thanks{*This work was partially performed during PH's internship at the Bosch Center for Artificial Intelligence.}
\thanks{$^{1}$PH, WD, BS, JF, DZ are with Carnegie Mellon University, USA. $^{2}$JF, BC are with Bosch Center for Artificial Intelligence. Contact: {\tt\footnotesize peideh@andrew.cmu.edu}}%
\vspace{-0.2cm}
}

\begin{document}

\maketitle
\thispagestyle{empty}
\pagestyle{empty}

\begin{abstract}
Simulation is an indispensable tool in the development and testing of autonomous vehicles (AVs), offering an efficient and safe alternative to road testing. An outstanding challenge with simulation-based testing is the generation of safety-critical scenarios, which are essential to ensure that AVs can handle rare but potentially fatal situations. This paper addresses this challenge by introducing a novel framework, \abbv, to generate realistic, diverse, and controllable safety-critical scenarios. Our approach optimizes for both the quality and diversity of scenarios by employing a unique formulation and algorithm that integrates real-world scenarios, domain knowledge, and black-box optimization. We validate the effectiveness of our framework through extensive testing in three representative types of traffic scenarios. The results demonstrate superior performance in generating diverse and high-quality scenarios with greater sample efficiency than existing reinforcement learning (RL) and sampling-based methods. 

\end{abstract}

\section{INTRODUCTION}

Simulation plays a pivotal role in the domain of autonomous driving, serving crucial functions in both training and evaluation of autonomous vehicles (AVs)~\cite{gulino2024waymax, herman2021learn, park2020diverse, xu2022trustworthy}. 
Compared to the costly and time-consuming on-road testing, simulation offers efficient feedback to developers and avoids risky engagements in the real world~\cite{huang2023went}.


It is widely recognized that traffic scenarios in the real world exhibit a long-tail distribution, where normal scenarios constitute the majority and safety-critical ones occur infrequently~\cite{ding2023survey, stoler2023safeshift}. AVs predominately trained on normal scenarios may not generalize to safety-critical ones, which can lead to fatal accidents when deployed at scale. 
Furthermore, evaluating AVs in such a long-tail distribution of naturalistic driving scenarios is neither sample-efficient nor comprehensive. Consequently, it is essential to generate safety-critical scenarios for both simulation-based training and evaluation.\looseness=-1 

There are three principal challenges in generating safety-critical scenarios. 
The first challenge is {\it realism}, that is, the scenarios generated should be sufficiently realistic to occur in the real world. This realism is often quantified by the similarity between real-world scenarios and generated ones~\cite{feng2023trafficgen, tan2021scenegen, huang2022curriculum}. To achieve this, existing methods either modify existing  scenarios~\cite{wang2021advsim, hanselmann2022king, stoler2024seal} or employ generative models to approximate the distribution of real-world scenarios and then modify the generated samples~\cite{zhong2023guided, tan2023language}. 
The second challenge is {\it diversity}, that is, the generation algorithm should cover a wide spectrum of scenarios rather than focusing on a single type of failure. 
Existing methods either lack in diversity or achieve it by construction, e.g., by specifying a multi-modal generative model \cite{ding2023survey}. 
The final challenge is {\it controllability}, that is the scenarios are generated according to specified characteristics.
These guidelines are often expressed through constraints~\cite{ding2021semantically}, temporal logic~\cite{zhong2023guided}, or language~\cite{tan2023language, xu2023creative}, all of which require the framework to be specially designed to facilitate their integration.\looseness=-1

To tackle these challenges, we reformulate safety-critical scenario generation within the Quality-Diversity(QD) framework and introduce our method \abbv. QD is a branch of optimization that finds a collection of high-performing yet qualitatively different solutions \cite{mouret2015illuminating, tjanaka2023pyribs}.   
Through a novel design, \abbv addresses the aforementioned challenges by integrating information from real-world data, domain knowledge, and black-box optimization.
To maintain \textit{realism}, we parameterize scenarios as bounded perturbations to real-world trajectories.
To promote \textit{diversity}, we improve on an existing QD algorithm and use it to efficiently explore and optimize the continuous parameter space.
To achieve \textit{controllability}, we retrieve archived scenarios according to the desired measure values.

\begin{figure*}[t!]
\begin{subfigure}[t]{0.66\textwidth}
  \centering
  \includegraphics[width=1\textwidth]{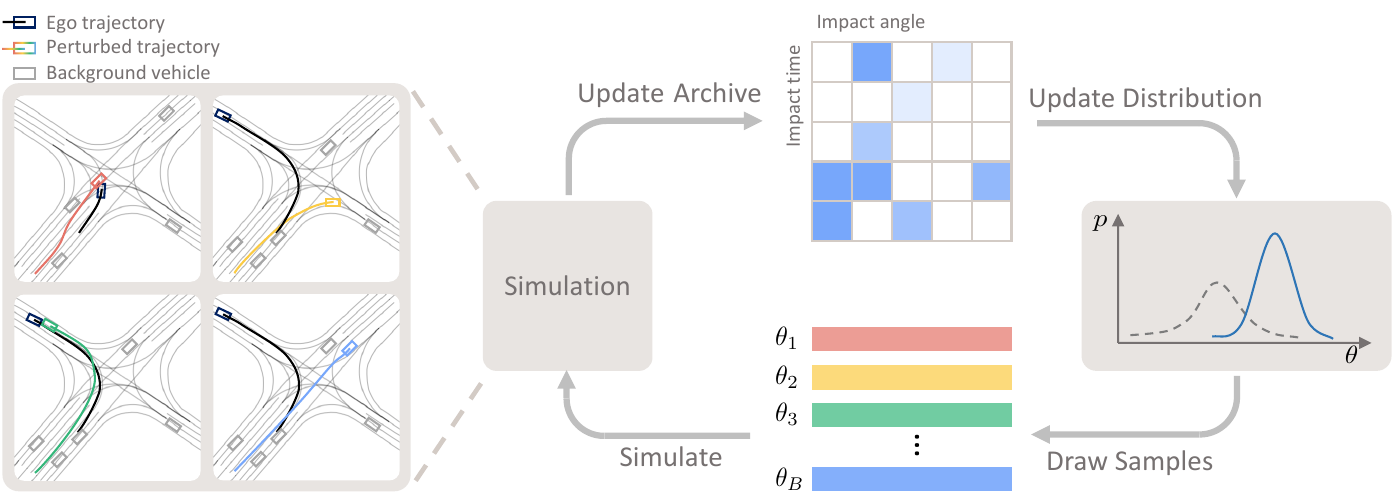}
  \caption{\;}
\end{subfigure}%
\hfill
\begin{subfigure}[t]{0.3\textwidth}
  \centering
  \includegraphics[width=1\textwidth]{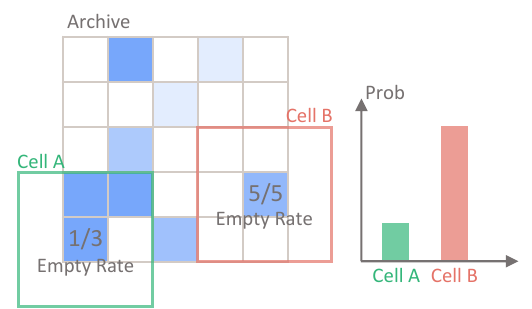}
  \caption{\;}
\end{subfigure}
\vspace{-0.2cm}
\caption{(a) Overview of the \abbv framework. \abbv utilizes a black-box optimization algorithm to explicitly optimize for high-quality and diverse safety-critical scenarios. (b) Illustration of the Occupancy-Aware Restart (OAR) mechanism, a general extension to QD algorithms to improve exploration efficiency.}
\vspace{-0.6cm}
\label{fig:overview}
\end{figure*}

\noindent The main contributions can be summarized below:
\begin{itemize}
    \item We propose \abbv, a novel QD formulation for safety-critical scenario generation, which achieves realism, diversity, and controllability
    .\looseness=-1 
    \item We propose an occupancy-aware restart (OAR) mechanism as a general extension to the QD algorithm family, which improves the exploration efficiency.
    \item We conduct experiments on three representative real-world traffic scenarios: unprotected cross-turn, high-speed lane-change, and U-turn, which demonstrate that \abbv generates diverse and high-quality scenarios with better sample efficiency compared to both RL- and sampling-based methods.
\end{itemize}



\section{RELATED WORK}

\para{Safety-critical Scenario Generation} 
One significant component of autonomous driving simulation is the traffic model, which governs the behavior of  background vehicles. Many works, e.g., TrafficSim~\cite{suo2021trafficsim}, TrafficGen~\cite{feng2023trafficgen}, ScenarioNet~\cite{li2023scenarionet}, focus on generating realistic scenarios in the naturalistic distribution.
In contrast, we focus on the long-tail of the distribution, that is the safety-critical scenarios, to provide efficient and comprehensive evaluations of the safety of AVs~\cite{ding2020cmts, ding2019multi, xu2022safebench}. 

A majority of such works focuses on adversarial generation.
L2C~\cite{ding2020learning}, MMG~\cite{ding2021multimodal}, and CausalAF~\cite{ding2021causalaf} generate initial conditions for open-loop scenario generation using RL.
The methods in \cite{wang2021advsim, klischat2019generating, arief2021deep, zhang2023cat} optimize the trajectories of actors with black-box optimization to attack the ego vehicle.
KING~\cite{hanselmann2022king} and AdvDO~\cite{cao2022advdo} further assume access to differential dynamics models to improve the efficiency of finding safety-critical scenarios.
However, adversarial attack-based methods do not explicitly optimize for the diversity and controllability of generated scenarios. To address this limitation, imitation learning~\cite{zhang2023learning}, retrieval-augmented generation~\cite{ding2023realgen}, causality~\cite{ding2022generalizing}, and evolutionary algorithms~\cite{li2021scegene} have also been explored.
To control the scenario generation with language, LCTGen~\cite{tan2023language} predefines an intermediate representation to bridge the large language model (LLM)~\cite{achiam2023gpt} and the low-level trajectory generator, and CTG++~\cite{zhong2023language} uses LLMs to generate signal temporal logic to guide the sampling process of diffusion models. 
In this paper, we depart from the common practice of leveraging adversarial generation methods and instead focus on improving the diversity and controllability of generated samples through our novel application and extensions of Quality-Diversity algorithms.\looseness=-1

\para{Quality-Diversity Algorithms in Robotics} QD is a branch of optimization that finds a collection of high-performing, yet qualitatively different solutions \cite{mouret2015illuminating, tjanaka2023pyribs}. 
Specifically, QD optimizes an objective for each point in a measure space. Solving a QD problem in a continuous measure space requires infinite memory \cite{tjanaka2023pyribs}. Thus, in practice, the measure space is discretized into a finite set, and an archive is maintained to keep track of the best-known solutions over the finite set. Some examples of popular algorithms include MAP-Elites \cite{mouret2015illuminating}, CMA-ME \cite{fontaine2020covariance}, and DQD \cite{fontaine2021differentiable}.

Given QD's ability to find a collection of high-performing solutions for different contexts, it is well-suited for many robotics applications. In \cite{cully2015robots}, QD is used to learn a behavior-performance map, enabling the robot to quickly find a compensatory behavior and adapt after damage. QD has also been used on problems such as human-robot interaction \cite{bhatt2023surrogate,tung2024workspace}, robot manipulation \cite{morel2022automatic, huber2024speeding}, locomotion \cite{surana2023efficient,nordmoen2021map}, and morphology design \cite{zardini2021seeking, Liu_2022}. To the best of our knowledge, we are the first to formulate QD for safety-critical scenario generation in autonomous driving; our approach enables us to generate an archive of diverse and high-quality scenes, and allows for fine-granular control by retrieving scenarios with desired characteristics as defined by the measure space.

\section{METHODOLOGY}


Our method, \textbf{C}ontrollable \textbf{a}nd \textbf{D}iverse Generation of Safety-Critical Driving Scenarios using \textbf{RE}al-world trajectories (\abbv), integrates real-world data, domain knowledge, and black-box optimization. As illustrated in Fig.~\ref{fig:overview}, for each iteration, \abbv maintains a grid archive of generated scenarios. First, it uses the QD algorithm to update the distribution from which the perturbations to the real-world trajectories are sampled. Then \abbv simulates the perturbations to obtain diverse behavior measures and updates the archive according to the simulation results. Finally, we obtain an archive that contains thousands of critical scenarios, each with different behaviors according to the measure functions we defined using domain knowledge.\looseness=-1

\subsection{Representation of Traffic Scenarios}\label{sec:scenario_representation}
We utilize real-world trajectories and design the parameterization of traffic scenarios to retain \textit{realism}.  Let $\bm{x}_t^i \in \mathbb{R}^2, \psi_t^i \in[-\pi, \pi]$ and $v_t^i \in \mathbb{R}$ be the world-frame coordinate, orientation, and speed of the $i$-th vehicle agent at time $t$. The ego vehicle, with index $i=0$, is the vehicle for which we want to generate critical scenarios. We denote the state of the vehicle as $\bm{s}_t=\left\{\bm{x}_t^i, \psi_t^i, v_t^i\right\}_{i=0}^N$, where $N$ is the number of background agents. We define a specific traffic scenario as a sequence of these states $\mathcal{S}=\left\{\bm{s}_t\right\}_{t=0}^T$, where $T$ is a fixed time horizon. We initialize a specific scenario from a real-world dataset that contains only naturalistic driving scenarios.

\para{Safety-Critical Perturbation} In each generated scenario, we perturb the trajectory of a single background vehicle indexed by $i \in [1, \ldots, N]$, which is selected according to criteria to be described in Section \ref{sec:setup}. We first recover the action sequence $\left\{\bm{a}_t^i\right\}^{T-1}_{t=0}$ from $\left\{\bm{s}_t^i\right\}^{T}_{t=0}$, assuming a kinematic bicycle model 
$[\dot x, \dot y, \dot \psi, \dot v] = [v \cos (\psi), v \sin (\psi), v \tan (\psi) / L, a] $,
where $L$ is the wheelbase. Each action consists of acceleration and steering input, i.e., $\bm{a}_t^i := [a_t^i, \delta_t^i]$. 
A new trajectory can be generated by 1) applying a sequence of bounded perturbations $\left\{\Delta\bm{a}_t^i\right\}^{T-1}_{t=0}$ to the recovered action sequence, and 2) unrolling the kinematics model from $\bm{s}_0^i$ using the bicycle model. We then parameterize each safety-critical scenario with $\bm{\theta}=\left\{\Delta\bm{a}_0^i, \ldots, \Delta\bm{a}_{T-1}^i\right\} \in \mathbb{R}^{T \times 2}$ and denote the allowable space of bounded perturbation as $\Theta$.\looseness=-1


\subsection{Quality-Diversity Formulation for Scenario Generation}\label{sec:qd_formulation} 

Inspired by previous work~\cite{bhatt2023surrogate,tjanaka2023pyribs}, we formulate the problem of generating a diverse set of safety-critical driving scenarios as a QD problem. First, we define an objective function $f: \mathbb{R}^{T \times 2}\rightarrow \mathbb{R}$ to quantify the safety-critical level. We further define K measure functions $m_k: \mathbb{R}^{T \times 2} \rightarrow \mathbb{R}$ and jointly represent them as $\bm{m}: \mathbb{R}^{T \times 2} \rightarrow \mathbb{R}^K$, which are a set of user-defined functions to quantify aspects of the scenario that we aim to diversify. We denote $\mathcal{M}= \bm{m}(\mathbb{R}^{T \times 2})\subseteq\mathbb{R}^K$ as the measure space formed by the range of $\bm{m}$. Because $f$ evaluates the quality of a scenario $\bm{\theta}$, the goal of the QD problem is to find, for each $m\in \mathcal{M}$, a scenario $\bm{\theta}$, such that $\bm{m}(\bm{\theta}) = m$ and that $f(\bm{\theta})$ is maximized (Eqn. \ref{eq:QD_continuous}):\looseness=-1
\vspace{-1mm}
\begin{equation}\label{eq:QD_continuous}
\begin{aligned}
    \max \quad &f(\bm{\theta})\\
    \text{s.t.}\quad & \bm{m}(\bm{\theta}) = m, \;\forall m\in \mathcal{M}.
\end{aligned}
\vspace{-1mm}
\end{equation}
In practice, we discretize $\mathcal{M}$ into a finite number of $M$ cells, which we will refer to as the archive, and solve the simplified version of the problem:
\vspace{-1.5mm}
\begin{equation}\label{eq:QD_discrete}
\max_{\bm{\theta}_1, \ldots, \bm{\theta}_M}\sum_{n=1}^M f(\bm{\theta}_n).
\end{equation}
\vspace{-2.5mm}

With a slight overload of notation, we also use $f$ to denote the objective value and $\bm m$ to denote the values of the measure function. We denote the archive as $M$, and we can retrieve the scenarios from the archive by $M[\bm m]$. With properly defined objective and measure functions, we can optimize a diverse population of safety-critical scenarios and retrieve individual scenarios in a controllable manner by asking for specific measure values $\bm m$. We build a lightweight traffic scenario simulator $Sim(\mathcal{S}, \bm\theta, i)$, which outputs the objective value $f$ and the measure values $\bm m$, given the original scenario $\mathcal{S}$, perturbation $\bm\theta$, and the index of the perturbed vehicle $i$.

For this setting, the typical formulation of safety-critical scenario generation would have been $\max_{\theta\in \Theta} f(\theta)$. To promote diversity, existing methods either parameterize the scenarios as a multi-modal distribution or impose an additional distributional constraint/regularization. In comparison, the QD formulation incorporates diversity directly as part of the problem.\looseness=-1 

\begin{algorithm}[t]
\SetAlgoLined
\SetKwProg{Fn}{Function}{:}{}
\textbf{Input: } Real-world normal scenario $\mathcal{S}$, index of the perturbed background vehicle $i$, traffic simulator $Sim$, batch size $B$, empty grid archive $M$ \\
\textbf{Output: } An grid archive $M$ containing diverse safety-critical scenarios $\mathcal{S}_c$ \\
Initialize emitter $e$ \\
Recover $\left\{\bm{a}_t^i\right\}^{T-1}_{t=0}$ from $\left\{\bm{s}_t^i\right\}^{T}_{t=0}$ in $\mathcal{S}$ \\
\For{iter = 1, $\ldots$, total\_iter}
{
$\{\bm \theta_b\}^B_{b=1} \sim \mathcal{N}(e.\mu, e.C)$ \\
    \For{$b = 1, \ldots, B$}
    {\{$f_b, \bm m_b\} \gets Sim(\mathcal{S}, \bm \theta_b, i)$}
Unpack $parents$, sampling mean $\mu$, covariance matrix $C$, and parameter set $P$ from e. \\
    \For{$b = 1, \ldots, B$}
    {
    \uIf{$M[\bm m_b]$ is empty}
    {
    $\Delta_b \gets f_b$ \\
    Flag that $\bm \theta$ discovered a new cell \\
    Add $\bm \theta_b$ to $parents$
    }
    \ElseIf{$f_b > M[\bm m_b].f$}
    {
    $\Delta_b \gets f_b - M[\bm m_b].f$ \\
    Add $\bm \theta_b$ to $parents$
    }
    }
    \eIf{parents $\neq \varnothing$}
    {
    Sort $parents$ by (newCell, $\Delta_b$) \\
    Update $\mu, C, P$ according to $parents$ \\
    $parents \gets \varnothing$ \\
    }
    {Occupancy-aware restart from an elite in $M$}
}
\caption{\abbv: Controllable and Diverse Generation of Safety-Critical Driving Scenarios}\label{alg:main}
\end{algorithm}

We specify the objective and measure functions here to demonstrate how concretely QD can be used to generate realistic, diverse, and controllable safety-critical scenarios. There could be many other alternative objective and measure function designs, and we hope other researchers will be inspired to adopt \abbv for their problems.\looseness=-1 

\para{Objective Function} The objective function $f$ quantifies the safety-critical level, motivated by prior work on safety-critical scenario generation (where $d(\cdot, \cdot)$ is the $l_2$ distance):
\vspace{-2mm}
\begin{equation}\label{eq:QD_objective}
f(\bm \theta) := \begin{cases}
      1, \quad \text{if vehicle $i$ collides with the ego vehicle}\\
      0, \quad \text{if vehicle $i$ collides with background vehicles} \\
      \exp(-\min_t d(\bm x_t^{0}, \bm x_t^{i})), \quad \text{otherwise} \\
    \end{cases} 
\end{equation}%
\para{Measure Functions} The measure functions are essential to capture different aspects of critical scenarios. We propose three measure functions to define the diverse behavior of perturbed vehicles. These measure functions collectively enable the definition and evaluation of a wide range of safety-critical scenarios, focusing on representative factors such as perturbation efforts ($m_1$), urgency of response ($m_2$), and collision behavior ($m_3$). 
Here, $m_1$ measures the mean magnitude of the steering perturbation. It partially reflects how much the generated trajectory would deviate from the original trajectory: %
\begin{equation}\label{eq:m1}
m_1 = \frac{1}{t_{\text{impact}}}\sum_{t=0}^{t_{\text{impact}}-1} \left|\delta_t^i\right|,
\end{equation}
\noindent Next, $m_2$ measures the normalized impact time; it helps categorize scenarios based on the urgency of the response required. When there is no collision, we assume $t_{\text{impact}} = \arg \min_t d(\bm x_t^{0}, \bm x_t^{i})$. This measure aids in the development of time-critical decision-making algorithms for AVs:
\begin{equation}\label{eq:m2}
m_2 = t_{\text{impact}}/T,
\end{equation}
\noindent Finally, $m_3$ measures the impact angle relative to the body frame of the ego vehicle. It allows for the evaluation of how well autonomous driving systems can recognize and react to threats from various directions, enhancing their ability to prevent accidents through appropriate maneuvering or braking:\looseness=-1
\begin{equation}\label{eq:m3}
m_3 = atan2(R_{\psi_t^0}(\bm{x}_t^i - \bm{x}_t^0)^T),
\end{equation}
where $R_{\psi_t^0}$ is the rotation matrix, $t=t_{\text{impact}}$, and $atan2$ is the 2-argument arctangent function. The $x$-axis of the body frame points to the vehicle's front, and the $y$-axis points to the left.

\subsection{\abbv Algorithm}
We summarize the algorithm of \abbv in Algorithm \ref{alg:main}. 
We modify a QD algorithm, namely Covariance Matrix Adaptation MAP-Elites (CMA-ME)~\cite{fontaine2020covariance}, by incorporating the domain-specific formulation (as described in Section \ref{sec:scenario_representation}, \ref{sec:qd_formulation}) and introducing OAR mechanism as a general extension to QD algorithms with discretized grid archives. 

\para{Algorithm Overview} The algorithm begins with the initialization of an archive and a population of solutions, which are then evaluated using $Sim$ and assigned to specific cells within the archive based on their measure values. The core of the algorithm involves iteratively selecting elite solutions according to 1) whether new cells are discovered (\textit{exploration}) and 2) how much the new solutions' objectives improve upon the old solution (\textit{exploitation}), which guides the adaptation of the CMA parameters, such as the mean $\mu$ and covariance $C$. These parameters shape the distribution from which new solutions are sampled. The algorithm proceeds by replacing less optimal solutions in the archive with better-performing ones, thereby refining the solution space exploration. This process continues until a predefined total iteration is reached, ultimately resulting in an archive that captures a diverse array of scenarios across the explored measure dimensions.


\para{Occupancy-Aware Restart} Existing QD algorithms~\cite{fontaine2020covariance} restart from random regions in the search space when there is no improvement in the archive. This is not efficient due to the possibility of restarting in well-explored regions. To improve exploration efficiency, we propose Occupancy-Aware Restart (OAR), a restart mechanism to serve as a form of guidance for improved coverage and efficiency during exploration. .


Illustrated in Fig. \ref{fig:overview}(b), OAR considers the occupancy rate of neighboring cells and assigns a higher probability to elites with more empty neighboring cells. More specifically, given the neighbor empty rate of $L$ elites $r_1, \ldots, r_L$ and the temperature $\tau$, the softmax probability of restarting from elite $i$ is computed by $p_i = \frac{e^{ r_i/\tau}}{\sum_{j=1}^{L} e^{r_j/\tau}}$.
OAR degenerates to the uniform sampling as $\tau\to\infty$. With a lower $\tau$, OAR assigns a higher probability to those elites who have more empty neighbors. For an efficient implementation, we use a 3D convolution kernel to compute the number of empty cells around each elite.

\begin{table*}[tb]
\centering
\caption{The final performance of coverage, mean objective, QD score. We report the mean and variance over 5 perturbed vehicles for each scene. The QD score is shown in multiples of $1000$. }
\vspace{-4mm}
\begin{center}
\begin{adjustbox}{width=1.0\linewidth}
\begin{tabular}{l|ccc|ccc|ccc}
\toprule 
\multirow{2}{*}{} & \multicolumn{3}{c}{Unprotected cross-turn} & \multicolumn{3}{c}{High-speed lane-change} & \multicolumn{3}{c}{U-turn} \\
Method    
& Coverage ($\uparrow$) & Mean Obj ($\uparrow$) & QD Score ($\uparrow$) 
& Coverage ($\uparrow$) & Mean Obj ($\uparrow$) & QD Score ($\uparrow$) 
& Coverage ($\uparrow$) & Mean Obj ($\uparrow$) & QD Score ($\uparrow$) \\
\midrule
Random       & 0.140$\pm$0.021 & 0.499$\pm$0.123 & 0.285$\pm$0.098 & 0.162$\pm$0.024 & 0.310$\pm$0.158 & 0.209$\pm$0.131 & 0.188$\pm$0.066 & 0.381$\pm$0.134 & 0.320$\pm$0.151\\
GOOSE        & 0.279$\pm$0.097 & 0.368$\pm$0.058 & 0.429$\pm$0.180 & 0.257$\pm$0.082 & 0.385$\pm$0.158 & 0.443$\pm$0.291 & 0.141$\pm$0.082 & 0.490$\pm$0.158 & 0.325$\pm$0.252\\
CMA-ES       & 0.182$\pm$0.033 & 0.672$\pm$0.076 & 0.489$\pm$0.090 & 0.163$\pm$0.018 & 0.540$\pm$0.143 & 0.347$\pm$0.086 & 0.228$\pm$0.118 & 0.447$\pm$0.228 & 0.502$\pm$0.253\\
REINFORCE    & 0.210$\pm$0.031 & 0.649$\pm$0.115 & 0.551$\pm$0.155 & 0.243$\pm$0.033 & 0.488$\pm$0.144 & 0.472$\pm$0.161 & 0.286$\pm$0.010 & 0.641$\pm$0.117 & 0.731$\pm$0.117\\
SVPG         & 0.219$\pm$0.032 & 0.607$\pm$0.130 & 0.553$\pm$0.155 & 0.265$\pm$0.026 & 0.410$\pm$0.129 & 0.437$\pm$0.166 & 0.290$\pm$0.020 & 0.577$\pm$0.120 & 0.665$\pm$0.110\\
\abbv (ours) & \textbf{0.565$\pm$0.054} & \textbf{0.829$\pm$0.062} & \textbf{1.884$\pm$0.309} & \textbf{0.541$\pm$0.079} & \textbf{0.627$\pm$0.142} & \textbf{1.375$\pm$0.436} & \textbf{0.542$\pm$0.094} & \textbf{0.793$\pm$0.073} & \textbf{1.699$\pm$0.219} \\
\bottomrule
\end{tabular}
\end{adjustbox}
\end{center}
\label{table:main}
\vspace{-6mm}
\end{table*}

\begin{figure*}[!t]
\begin{subfigure}{\textwidth}
  \centering
  \includegraphics[width=0.85\textwidth]{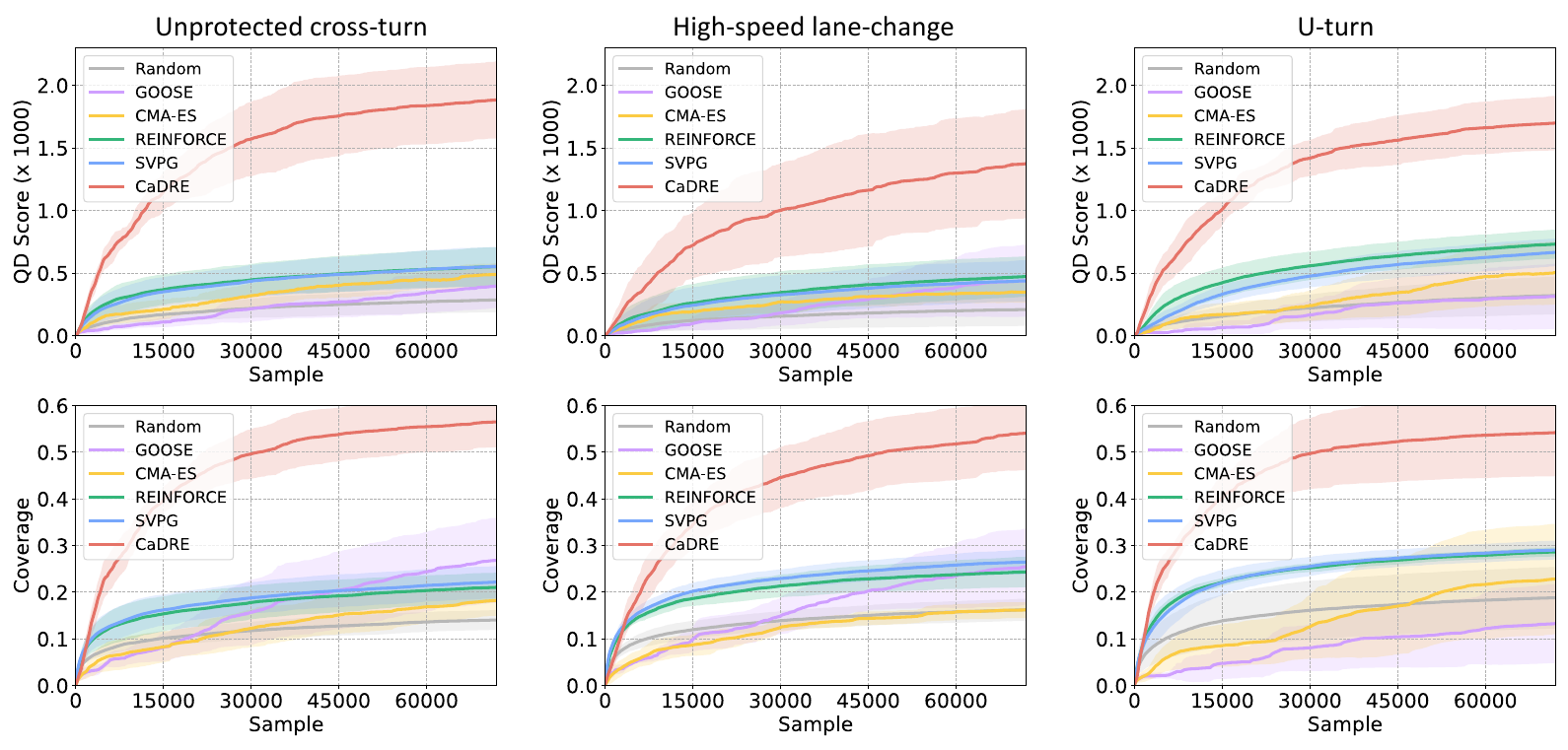}
\end{subfigure}
\vspace{-8mm}
\caption{Coverage and QD score v.s. number of samples. Solid lines represent the mean, and the shaded area presents the standard deviation over 5 perturbed vehicles. 
}
\label{fig:curve}
\vspace{-0.4cm}
\end{figure*}

\section{EXPERIMENTS}

\subsection{Experimental Setup}\label{sec:setup}
\para{Real-world Trajectories} We pick three representative scenarios from nuPlan v1.1 \cite{caesar2021nuplan}: \textit{unprotected cross-turn}, \textit{high-speed lane-change}, and \textit{U-turn}. All scenarios have a total time horizon of 15 seconds.

\para{Reactive Ego Policy} Following~\cite{hanselmann2022king, cao2022advdo}, we assume that the ego vehicle is reactive to nearby vehicles. We implement a rule-based ego policy: The ego vehicle will try to follow the reference trajectory, and if there is a vehicle within $5m$ and $[-\pi/4, \pi/4]$ of the body frame of the ego vehicle, it will brake at $-7m/s^2$ and steer away from that vehicle with maximum steering angle $\pm\pi/8$. Since the problem is formulated as a black-box optimization, \abbv does not require access to the reactive policy and would work with any other ego policies.

\para{Selection of Perturbed Vehicles and Perturbation Range} To ensure effective perturbation, we use a simple heuristic to select which vehicles to perturb: the top five background vehicles that have the smallest average distance to the ego vehicle. Acceleration perturbation is between $\pm 2$, and steering perturbation range is between $\pm \pi/8$.

\para{Evaluation Metrics} We focus on three criteria that measure the quality and diversity of the generated scenarios, which are standard metrics in the QD literature~\cite{fontaine2020covariance, tjanaka2023pyribs, fontaine2021differentiable, fontaine2020quality}.
\begin{itemize}
    \item \textbf{Coverage} $\in [0, 1]$: Proportion of cells in the archive that have an elite. This is a measure of \textit{diversity} of the generated scenarios. Note that while coverage is a standard metric for QD, there lacks such a principled metric to evaluate \textit{diversity} in safety-critical scenario generation literature. 
    \item \textbf{Mean objective} $\in [0, 1]$: Mean objective value of elites in the archive.
    \item \textbf{QD score} $\in [0, 4000]$: Sum of the objective values of all elites in the archive. The theoretical maximum value of $4000$ is due to our objective $f \in [0, 1]$, and we discretize the measure space into $10 \times 20 \times 20$ grid. 
\end{itemize}

\para{Baselines} We study a mixture of sampling- and RL-based methods that have been employed by existing literature.
\begin{itemize}
    \item Random Search (\textbf{Random}): uniformly-random sample from the solution space.
    \item \textbf{GOOSE} \cite{ransiek2024goose}: a state-of-the-art RL-based approach for safety-critical scenario generation. It parameterizes adversarial trajectories as non-uniform rational B-splines (NURBS) and iteratively modifies their control points for diverse scenario generation. Note that this is also an example of diversity by construction, where diversity is promoted through the representation of scenarios.  
    \item \textbf{CMA-ES} \cite{hansen2016cma}: CMA-ES is an evolutionary strategy that iteratively updates a population of solutions based on their fitness
    and adaptively adjusts the search distribution towards optimal regions of the solution space.
    \item Multi-particle \textbf{REINFORCE} \cite{williams1992simple}: policy gradient method employed by previous work \cite{ding2020learning, ding2021multimodal}. We set the number of particles to be the same as the batch size ($36$) of CMA-ME employed by our algorithm.
    \item Stein Variational Policy Gradient (\textbf{SVPG}) \cite{liu2017stein}: SVPG is an improved version of multi-particle REINFORCE by using a maximum entropy policy optimization framework that explicitly encourages diverse solutions and better exploration. The number of particles is the same as the batch size of \abbv.
\end{itemize}

We adopt the QD algorithm library \textit{pyribs} \cite{tjanaka2023pyribs} to implement our framework. We aim to answer the following questions in our experimental study:
\begin{itemize}
    \item How does \abbv compare with baseline methods in terms of the evaluation metrics and sample efficiency?
    \item Is OAR effective in improving exploration?
    \item Can we retrieve diverse scenarios generated by \abbv in a controllable manner?
\end{itemize}

\subsection{Sample-efficiency Compared with Baseline Methods}

\begin{figure*}[t]
\begin{subfigure}{\linewidth}
  \centering
  \includegraphics[width=0.99\textwidth]{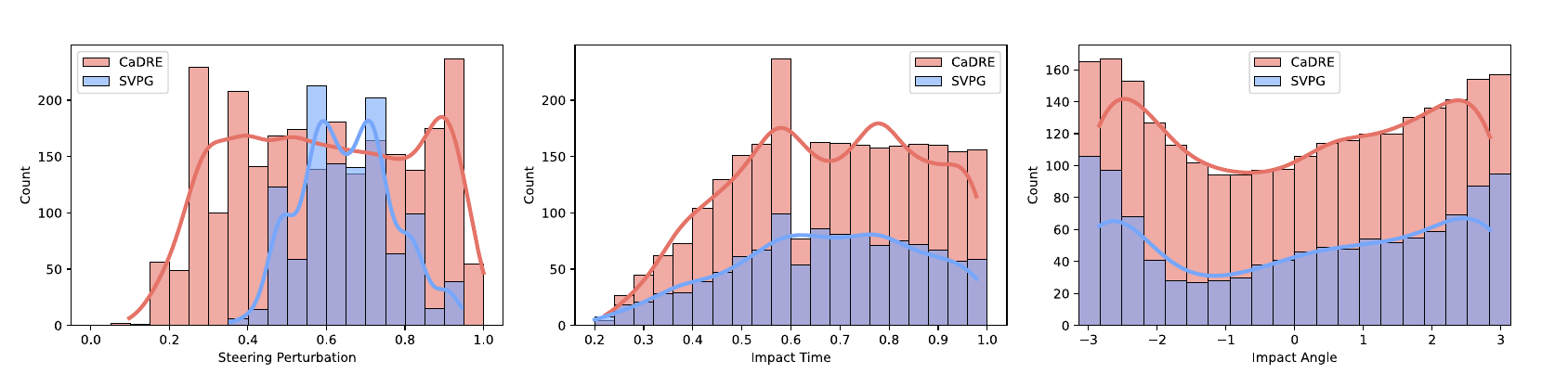}
\end{subfigure}
\vspace{-0.6cm}
\caption{Histograms of measure values. We visualize the final archive of the perturbed background vehicle with the highest QD score in the unprotected cross-turn. The solid line is the kernel density estimate of the true distribution. Note that the Gaussian kernel may introduce some distortions since the true distribution is bounded.}
\label{fig:archive_dist}
\vspace{-0.2cm}
\end{figure*}

The coverage and QD score v.s. samples are shown in Figure ~\ref{fig:curve}. \abbv outperforms all baselines with significant margins in three different scenarios, which demonstrates that \abbv discovers both high-quality and diverse scenarios more sample-efficiently than Random Search, GOOSE, SVPG, and REINFORCE. Under the hood, \abbv adapts the search distribution over generations to increase the likelihood of sampling in promising areas of the solution space, in contrast to Random Search, which samples uniformly across the solution space. CMA-ES maximizes the likelihood of increasing the objective and thus quickly converges to a single optimum. SVPG and REINFORCE, while exploring more than CMA-ES, still struggle with exploring complex problem spaces due to their focus on gradient-based optimization. Finally, GOOSE, a dedicated method for safety-critical scenario generation, claimed that the NURBS parameterization of traffic scenarios is sufficient for \textit{diversity}. However, it is clear from the results that its diversity, as measured by coverage, lags far behind \abbv, though being competitive with other baselines.

\begin{figure*}[!]
\begin{subfigure}{\textwidth}
  \centering
  \includegraphics[width=0.9\textwidth]{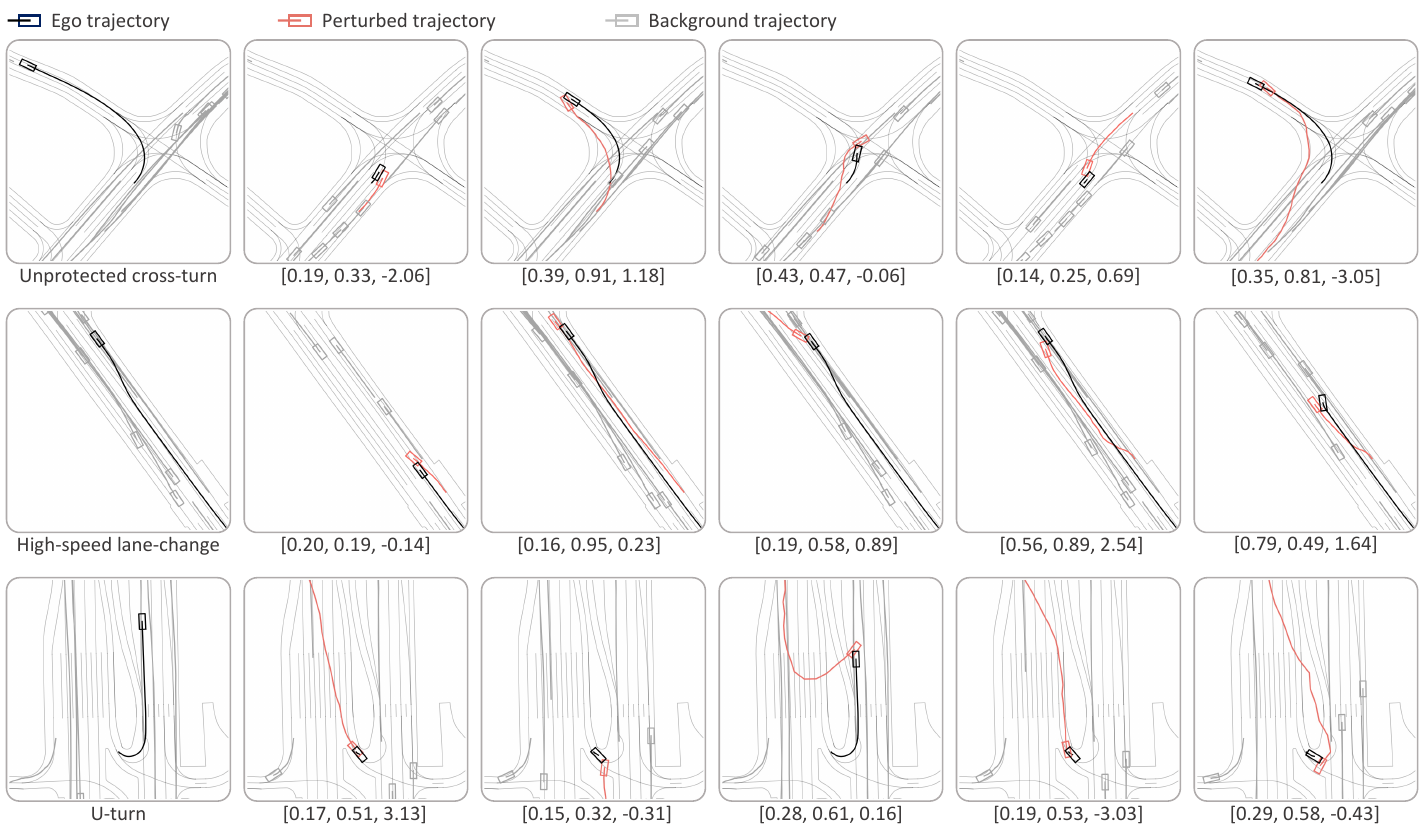}
\end{subfigure}
\vspace{-8mm}
\caption{Visualization of generated trajectories. The leftmost column shows the original unperturbed scenarios. The numbers below are the measure values $[m_1, m_2, m_3]$, representing the mean steering perturbation, impact time, and impact angle, respectively.}
\label{fig:traj_vis}
\vspace{-0.5cm}
\end{figure*}

\begin{figure}[!t]
\begin{subfigure}{\linewidth}
  \centering
  \includegraphics[width=0.85\textwidth]{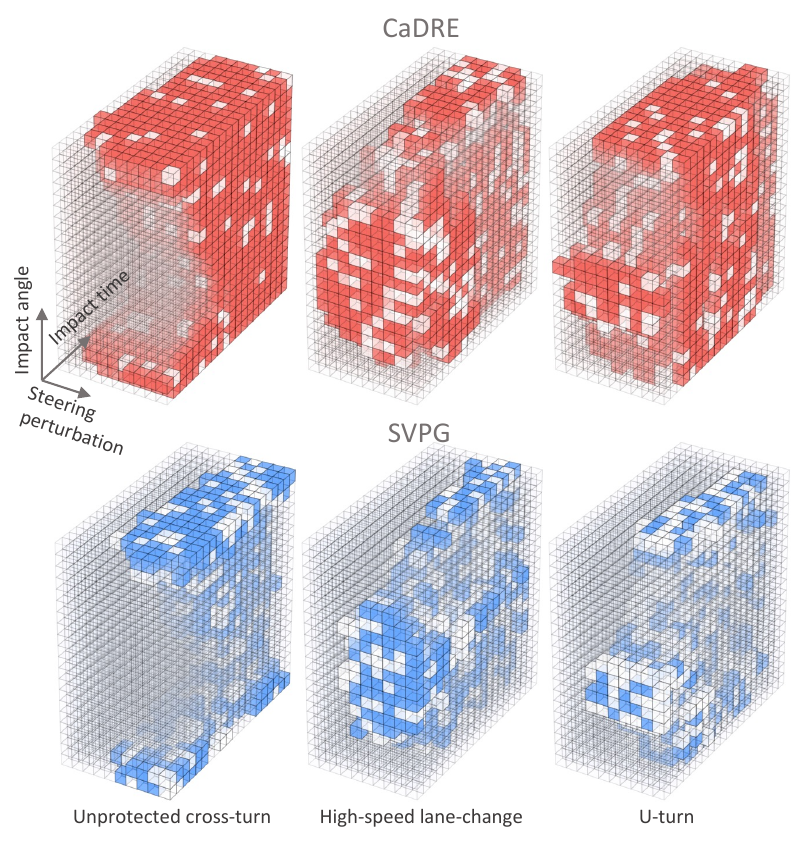}
\end{subfigure}
\vspace{-7mm}
\caption{Visualization of final archives. A darker color means a higher objective value. Transparent cells mean we cannot find scenarios. We visualize the perturbed vehicles that have the highest QD score for each scenario respectively.}
\label{fig:archive_vis}
\end{figure}

Table~\ref{table:main} shows the final performance of coverage, mean objective, and QD score. With the same number of samples, \abbv achieves $158.0\%, 122.6\%, 89.5\%$ more coverage, $36.6\%, 28.5\%, 23.7\%$ higher mean objective, leading to a $240.7\%, 191.3\%, 132.4\%$ improvement in QD score than the best-performing baselines in three representative scenarios, respectively. It again highlights the superior exploration and exploitation capability of \abbv compared to the baselines.

Table~\ref{table:oar} shows the ablation of OAR in the high-speed lane-change scenario. $\tau\to\infty$ is equivalent to QD algorithms without OAR. OAR improves the QD score of individual vehicles by a maximum of $33.0\%$, which demonstrates the effectiveness of OAR. We also observe that the biggest improvements come from harder problems (low QD score). The intuition is that when it is harder to find a safety-critical scenario, then a more effective exploration strategy provides a more significant performance boost.  

\begin{table}[!t]
\centering
\caption{QD score of occupancy-aware restart with different temperatures. The QD score is shown in multiples of $1000$. The percentage improvements w.r.t $\tau\to\infty$ are in parentheses.}
\vspace{-3mm}
\begin{center}
\begin{adjustbox}{width=1.0\linewidth}
\begin{tabular}{c|lllll}
\toprule 
Index & $1$ & $2$ & $3$ & $4$ & $5$\\
\midrule
 $\tau\to\infty$     & 1.808           & 0.519          &  \textbf{1.277} & 1.393 & 1.444 \\
 $\tau =1/ 5$     & 1.855 (2.6\%)   & 0.639 (22.9\%) & 1.133 (-11.3\%) & \textbf{1.515 (8.8\%)} & 1.299 (-10.0\%) \\
 $\tau = 1/10$     & \textbf{2.026 (12.0\%)}  & \textbf{0.691 (33.0\%)}        & 1.229 (-3.7\%) & 1.367 (-1.9\%) & \textbf{1.563 (8.2\%)} \\
\bottomrule
\end{tabular}
\end{adjustbox}
\end{center}
\label{table:oar}
\end{table}

\subsection{Analysis of the Generated Safety-Critical Scenarios}

\para{Visualization of Archives} We visualize the final archives in Fig.~\ref{fig:archive_vis}. It is observed that the proposed \abbv leads to a much higher occupancy as well as mean objective in the final archives than the best-performing baseline SVPG. The main reason is that \abbv explicitly encourages sustained exploration throughout optimization. 
Although the repulsive force in SVPG indeed introduces diversity among the particles to avoid premature convergence to local optima, the primary focus remains on optimizing a solution rather than explicitly seeking out diverse solutions across a range of measures.

Note that some cells are still unoccupied even for \abbv. We hypothesize that it is due to the infeasibility of finding scenarios, which is induced by specific combinations of measure values, the vehicle states in the original scenarios, and the kinematics constraints. For example, it is extremely difficult to find a solution with a short impact time and a small impact angle (hitting from the front) in the unprotected cross-turn scenario, since there is no background vehicle starting near the front of the ego vehicle.


\para{Visualization of Generated Scenarios} In Fig.~\ref{fig:traj_vis}, we visualize five generated scenarios for the unprotected cross-turn, high-speed lane-change, and U-turn, respectively. The visualization shows that we are able to generate and retrieve diverse critical scenarios in a controllable manner. For example, in the unprotected cross-turn, we can control the perturbed vehicle hitting the right side of the ego vehicle by steering a little bit from the original trajectory or hitting the left side by overtaking from the left, simply by asking for different combinations of measure function values $[m_1, m_2, m_3]$ in the archive.

\section{CONCLUSIONS AND DISCUSSION}
In this work, we develop a framework \abbv for generating safety-critical scenarios. \abbv enhances the diversity and controllability of the scenario generation process while retaining realism. We conduct extensive experiments on three representative scenarios: unprotected cross-turn, high-speed lane-change, and U-turn. The experimental results show that \abbv can generate and retrieve diverse and high-quality scenarios with better sample efficiency compared to existing methods.\looseness=-1

We envision many extensions possible to \abbv. At the moment, \abbv only perturbs one background vehicles at a time. Alternative parameterization of traffic scenarios or objective and measure functions can make \abbv more general or suited to specific settings. Second, \abbv does not consider the lane information and road conditions such as barriers. It is possible to incorporate road information to \abbv by augmenting the existing simulator or adding a post-hoc pruning step based on the drivable area.







\balance
\bibliographystyle{IEEEtran} 
\bibliography{citation}

\end{document}